\documentclass[10pt,twocolumn,letterpaper]{article}

\usepackage[pagenumbers]{cvpr} % To force page numbers, e.g. for an arXiv version

\usepackage[dvipsnames]{xcolor}

\definecolor{cvprblue}{rgb}{0.21,0.49,0.74}
\usepackage[pagebackref,breaklinks,colorlinks,citecolor=cvprblue]{hyperref}

\def\paperID{****} % *** Enter the Paper ID here
\def\confName{CVPR}
\def\confYear{2024}

\usepackage[dvipsnames]{xcolor}
\usepackage{color}
\usepackage{colortbl}
\usepackage{multirow}
\usepackage{tikz}  %%% change file in input in eso-pic.sty
\usepackage{pgfplots}
\usetikzlibrary{fit}
\pgfplotsset{compat=newest}%
\def\modelname{SED\xspace}
\def\R{{\mathbb R}}
\def\sta{FAM\xspace}
\def\ffu{SFM\xspace}
\newcommand{\hlrow}{\rowcolor{black!6}}

\title{SED: A Simple Encoder-Decoder for Open-Vocabulary Semantic Segmentation}

\author{Bin Xie$^{1}$, Jiale Cao$^{1}$, Jin Xie$^2$,   Fahad Shahbaz Khan$^3$, and Yanwei Pang$^{1,4}$\\
$^1$Tianjin University~~~$^2$Chongqing University\\$^3$Mohamed bin Zayed University of Artificial Intelligence\\
$^4$Shanghai Artificial Intelligence Laboratory\\
{\tt\small \{bin\_xie,connor,pyw\}@tju.edu.cn} 
{\tt\small xiejin@cqu.edu.cn}~ {\tt\small fahad.khan@mbzuai.ac.ae}
}

\begin{document}
\maketitle

\begin{abstract}
Open-vocabulary semantic segmentation strives to distinguish pixels into different semantic groups from an open set of categories. Most existing  methods explore utilizing pre-trained vision-language models, in which the key is to adopt the image-level model for pixel-level segmentation task. In this paper, we propose a simple encoder-decoder, named SED, for open-vocabulary semantic segmentation, which comprises a hierarchical encoder-based cost map generation and a gradual fusion decoder with category early rejection. The hierarchical encoder-based cost map generation employs hierarchical backbone, instead of plain transformer, to predict pixel-level image-text cost map. Compared to plain transformer, hierarchical backbone better captures local spatial information and has linear computational complexity with respect to input size. Our gradual fusion decoder employs a top-down structure to combine cost map and the feature maps of different backbone levels  for segmentation. To accelerate  inference speed, we introduce a category early rejection scheme in the decoder that rejects many no-existing categories at the early layer of decoder, resulting in at most  4.7 times acceleration without accuracy degradation. Experiments are performed on multiple open-vocabulary semantic segmentation datasets, which demonstrates the efficacy of our SED method. When using ConvNeXt-B, our  SED method achieves mIoU score of 31.6\% on ADE20K with 150 categories at 82 millisecond ($ms$) per image on a single A6000. We will release it at \url{https://github.com/xb534/SED.git}.
\end{abstract}    
\section{Introduction}
\label{sec:intro}
Semantic segmentation is one of the fundamental computer vision tasks that aims to parse the semantic categories of each pixel in an image. Traditional semantic segmentation methods~\cite{long2015fully,chen2017deeplab,xie2021segformer} assume that the semantic categories are closed-set and struggle to recognize the unseen semantic category during inference. To this end, recent works have explored open-vocabulary semantic segmentation~\cite{zhao2017open, SPNet,ZS3Net} that aims  to segment the pixels belonging to the arbitrary semantic categories.

Recently, vision-language models, such as CLIP~\cite{CLIP} and ALIGN~\cite{align}, learn aligned image-text feature representation from millions of  image-text paired data. The pre-trained vision-language models exhibit superior generalization ability to recognize open-vocabulary  categories. This motivates a body of research works to explore using vision-language models for open-vocabulary semantic segmentation~\cite{zegformer,ovseg}. Initially, research works mainly adopt two-stage framework~\cite{xu2022simple,ovseg,maskclip} to directly adapt vision-language models for open-vocabulary  segmentation. Specifically, they first  generate class-agnostic mask proposals and then adopt the pre-trained vision-language models to classify these proposals into different categories. However, such a two-stage framework uses two independent networks for mask generation and classification, thereby hampering computational efficiency. Further, it does not fully utilize the contextual information. 

\begin{figure}[t]
\centering
\includegraphics[width=1.0\linewidth]{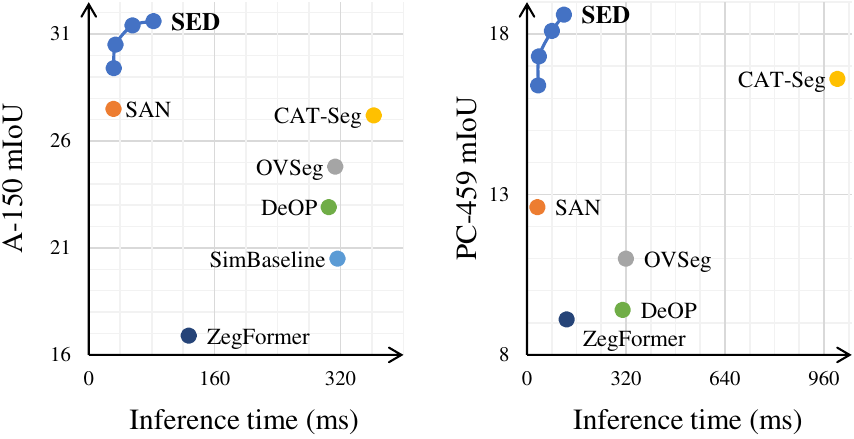}
\vspace{-10pt}
\caption{\textbf{Accuracy (mIoU) and speed (ms) comparison} on \texttt{A-150}  and \texttt{PC-459}.  Here, the speed is reported on a single NVIDIA A6000 GPU.  Our proposed SED achieves an optimal trade-off in terms of speed and accuracy compared to existing methods in literature: SAN \cite{san}, CAT-Seg \cite{catseg}, OVSeg \cite{ovseg}, DeOP \cite{Han2023ZeroShotSS}, SimBaseline \cite{xu2022simple} and ZegFormer \cite{zegformer}.}
\vspace{-10pt}
\label{fig:intro}
\end{figure}

Different to the aforementioned two-stage approaches, methods based on the single-stage framework directly extend a single vision-language model for open-vocabulary segmentation. Several methods remove the pooling operation in last layer of image encoder and generate pixel-level feature map for segmentation. For instance, MaskCLIP~\cite{maskclip1} removes the global pooling at last layer of the CLIP image encoder and uses the value-embeddings and text-embeddings to directly predict pixel-level segmentation map. CAT-Seg~\cite{catseg} first generates pixel-level image-text cost  map and then refines the cost map with spatial and class aggregation. While these approaches achieve favorable performance compared to their two-stage counterparts, we note their following limitations.
First, both MaskCLIP~\cite{maskclip1} and CAT-Seg employ plain transformer ViT~\cite{vit} as the backbone which suffers from weak local spatial information and low-resolution input size. To address those issues, CAT-Seg introduces an additional network to provide spatial information. However, this incurs  extra computational cost. Second, the computational cost of CAT-Seg significantly increases with the larger number of open-vocabulary classes. 

To address the aforementioned issues, we propose a simple yet effective encoder-decoder approach, named SED, for open-vocabulary semantic segmentation. Our proposed SED comprises a hierarchical encoder-based cost map generation and a gradual fusion decoder with category early rejection.  The hierarchical encoder-based cost map generation employs hierarchical backbone, instead of plain transformer, to predict pixel-level image-text cost map. Compared to plain transformer, hierarchical backbone better preserves the spatial information at different levels and has a linear computational complexity with respect to the input size. Our gradual fusion decoder gradually combines the feature maps from different levels of hierarchical backbone and cost map for segmentation prediction. To increase the inference speed, we design a category early rejection scheme  in the decoder that effectively predicts existing categories and rejects non-existing categories at the early layer of the decoder. Comprehensive experiments are conducted on multiple open-vocabulary semantic segmentation datasets, revealing the merits of the proposed contributions in terms of accuracy and efficiency. To summarize, we propose a simple yet effective open-vocabulary semantic segmentation approach with the following contributions.

\begin{itemize}
    \item We propose an encoder-decoder for open-vocabulary semantic segmentation comprising a hierarchical encoder-based cost map generation and a gradual fusion decoder.

    \item We introduce a category early rejection scheme to reject non-existing categories at the early layer, which aids in markedly increasing the inference speed without any significant degradation in  segmentation performance. For instance, it provides 4.7  times acceleration on PC-459.
    
    \item Our proposed method, SED, achieves the superior performance on  multiple open-vocabulary  segmentation datasets. Specifically, the proposed SED provides a good trade-off in terms of segmentation performance and speed (see Fig. \ref{fig:intro}). When using ConvNeXt-L, our proposed SED obtains mIoU scores of 35.2\% on \texttt{A-150} and 22.6\% on \texttt{PC-459}.
    
\end{itemize}

\begin{figure*}[t]
\centering
\includegraphics[width=0.88\linewidth]{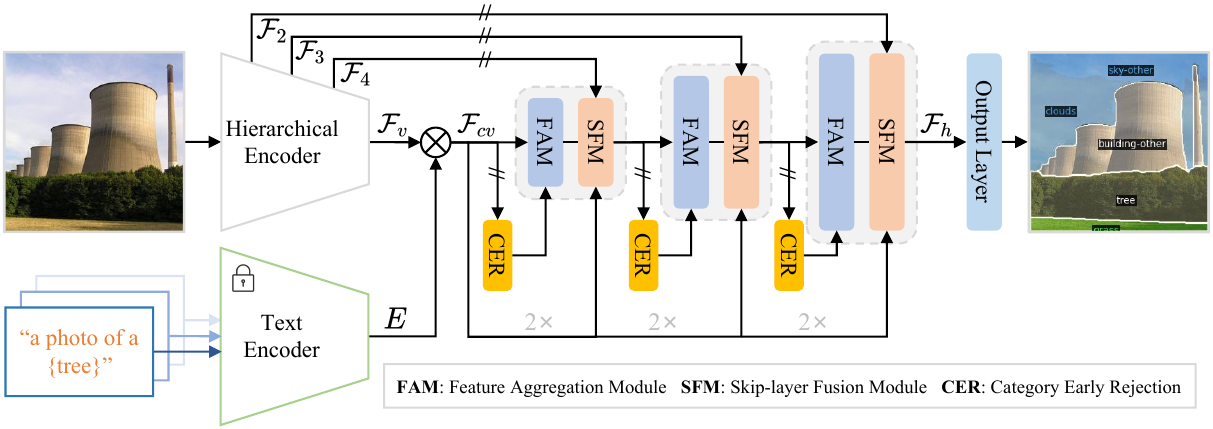}
\vspace{-10pt}
\caption{\textbf{Overall architecture of our proposed \modelname.} We first employ hierarchical  encoder (learnable) and text encoder (frozen) to generate pixel-level image-text cost map. Afterwards, we introduce a gradual fusion decoder to combine different feature maps of hierarchical encoder and cost map. The gradual fusion decoder stacks feature aggregation module (FAM) and skip-layer fusion module (SFM). In addition, we design a category early rejection (CER) in decoder to accelerate inference speed without sacrificing  performance.}
\label{fig:main_arch}
\vspace{-10pt}
\end{figure*}
\section{Related Work}
\label{sec:relate_work}

\subsection{Semantic Segmentation}
Traditional semantic segmentation methods mainly contain FCN-based approaches and transformer-based approaches. 
 Initially, the researchers focused on  FCN-based approaches. Long \textit{et al.} \cite{long2015fully} proposed one of the earliest fully-convolutional networks that fuses both deep and shallow features for improved segmentation. Afterwards, many FCN-based variants were proposed. Some methods utilize  spatial pyramid network ~\cite{chen2017deeplab,yang2018denseaspp} and encoder-decoder structure ~\cite{badrinarayanan2017segnet,tripleNet} to extract local contextual information. Some  methods \cite{danet,huang2019ccnet} exploit attention module to extract non-local contextual information. Recently, the researchers focused on developing transformer-based approaches. Some methods~\cite{xie2021segformer,gu2022multi,chen2022vision,lu2023content} employ the transformer as backbone to extract deep features, while some methods~\cite{setr,strudel2021segmenter,maskformer,cheng2022masked} treat transformer  as a segmentation decoder.

\subsection{Vision-Language Models}
Vision-language models aims to learn the connection between image representation and text embeddings. Initially, the researchers developed the vision-language models~\cite{Vilbert,tan2019lxmert,chen2020uniter} based on the pre-trained visual and language models, and explored to  jointly fine-tune them on different downstream tasks with image-text pairs. In contrast, CLIP \cite{CLIP} collects a large-scale image-text paired data from website and learns visual features via language supervision from scratch. The learned CLIP on large-scale data has a superior performance on different zero-shot tasks. Instead of using cleaned image-text paired data, ALIGN \cite{align} learns visual-language representation from noisy image-text dataset. To achieve this goal, ALIGN employs a dual-encoder structure  with contrastive loss, which achieves a good zero-shot performance on downstream tasks. Recently, Cherti \textit{et al.} ~\cite{cherti2023reproducible} conducted deep analysis on contrastive language-vision learning. Schuhmann \textit{et al.}~\cite{schuhmann2022laion} built a  billion image-text paired dataset for training large-scale  vision-language models.

\subsection{Open-Vocabulary Semantic Segmentation}
Open-vocabulary semantic segmentation aims at segmenting arbitrary categories. Initially, the researchers~\cite{zhao2017open, SPNet,ZS3Net} explored to align  visual features with  pre-trained text embeddings via a learned feature mapping. With the success of  large-scale vision-language model CLIP~\cite{CLIP}, the researchers started to explore open-vocabulary semantic segmentation using CLIP. Some methods \cite{zegformer,xu2022simple} adopt two-stage framework that first predicts class-agnostic mask proposals and second classifies these proposals into different categories. To improve classification performance at second stage, OVSeg \cite{ovseg} fine-tunes the CLIP model  on the masked image and their text annotations.  Ding \textit{et al.}~\cite{maskclip} integrated mask tokens with pre-trained CLIP model for mask refinement and classification.  ODISE~\cite{odise}  employs text-to-image diffusion model to generate mask proposals and perform classification. To enhance open-vocabulary performance, ODISE~\cite{odise} further performs mask classification using the  features cropped from pre-trained CLIP. 

In contrast, some methods adopt single-stage framework. 
LSeg~\cite{lseg} learns pixel-level image features guided by the  pre-trained CLIP text embeddings. MaskCLIP~\cite{maskclip1} removes the self-attention pooling layer to generate pixel-level feature map and employs text-embeddings to predict final segmentation map. SAN~\cite{san} introduces a side adapter network along the frozen CLIP model to perform mask prediction and classification. FC-CLIP \cite{fcclip} employs a frozen convolutional CLIP to predict class-agnostic masks and employs mask-pooled features for classification. CAT-Seg \cite{catseg} generates pixel-level cost map and refines the cost map for segmentation prediction. 
Our proposed method is inspired by CAT-Seg that  fine-tuning image encoder through cost map does not  degrade its open-vocabulary ability, but has significant differences: (1) Our  SED is a simpler framework without additional backbone, and has a better performance and faster inference speed. (2) Our SED employs hierarchical image encoder to generate cost map and to perform skip-layer fusion, which can significantly improve performance and has linear computational cost with respect to input size. (3) In decoder, we introduce a simple large-kernel operation and gradual fusion  for feature aggregation, and design a category early rejection strategy for acceleration without sacrificing performance.

\section{Method}
In this section, we describe our proposed encoder-decoder for open-vocabulary semantic segmentation, named \modelname. Fig. \ref{fig:main_arch} shows the overall architecture of our proposed \modelname, which comprises  two main components: a hierarchical encoder-based cost map generation and a  gradual fusion decoder with category early rejection. In our hierarchical encoder-based cost map generation, we employ hierarchical image encoder and text encoder to generate pixel-level image-text cost map $\mathcal{F}_{cv}$ and hierarchical feature maps $\mathcal{F}_{2},\mathcal{F}_{3},\mathcal{F}_{4}$ for  the decoder. Our gradual fusion decoder employs feature aggregation module (FAM) and skip-layer fusion module (SFM) to gradually combine pixel-level cost map $\mathcal{F}_{cv}$ and hierarchical feature maps $\mathcal{F}_{2},\mathcal{F}_{3},\mathcal{F}_{4}$  for generating high-resolution feature map $\mathcal{F}_{h}$. Based on $\mathcal{F}_{h}$, we employ an output layer to predict segmentation maps of different categories. In addition, a category early rejection (CER) strategy is used in the decoder to early reject non-existing categories for boosting inference speed.

\subsection{Hierarchical Encoder-based Cost Map}
Hierarchical encoder-based cost map generation (HECG) adopts the vision-language model CLIP~\cite{schuhmann2022laion,cherti2023reproducible,CLIP} to  generate pixel-level image-text cost map. Specifically, we first employ hierarchical image encoder and a text encoder to respectively extract  visual features and text embeddings. Then,  we calculate pixel-level cost map between these two features.  Existing methods such as MaskCLIP \cite{maskclip1} and CAT-Seg \cite{catseg} adopt the plain transformer as image encoder to generate pixel-level cost map. As discussed earlier, plain transformer suffers from relatively weak local spatial information and has quadratic complexity with respect to the input size. To address those issues, we propose to use hierarchical backbone as image encoder for cost map generation. Hierarchical encoder can better capture local information and has linear complexity with respect to the input size. The cost map generation is described as follow.

Given an input image $I \in \R^{H \times W \times 3}$, we first utilize a hierarchical  encoder ConvNeXt~\cite{convnext,schuhmann2022laion} to extract multi-scale feature maps, denoted as $\mathcal{F}_2, \mathcal{F}_3, \mathcal{F}_4,\mathcal{F}_5$. These feature maps have strides of  $4, 8, 16, 32$ pixels with respect to the input size. To align the output visual features and text embeddings, an MLP layer is attached at the last feature map $\mathcal{F}_5$ to obtain an aligned visual feature map $\mathcal{F}_{v} \in \R^{H_{v} \times W_{v} \times D_{t}}$, where ${D_t}$ is equal to the feature dimension of text embeddings, $H_v$ is  $H/32$, and $W_v$ is $W/32$.
Given an arbitrary set of category names $\{T_1,..,T_N\}$, we use the prompt template strategy~\cite{catseg,ovseg,gu2021open}  to generate different textual descriptions $S(n) \in \R^{P}$ about  category name $T_n$,  such as ``a  photo of a $\{T_n\}$, a photo of many $\{T_n\}, ...$". $N$ represents the total number of categories, and $P$ is the number of templates for each category. By fed $S(n)$ to the  text encoder, we obtain text embeddings, denoted as $E=\{E_1,..,E_N\} \in \R^{N \times P \times D_t}$.
By calculating the cosine similarity~\cite{rocco2017convolutional} between visual feature map $\mathcal{F}_{v}$ and text embeddings $E$, we obtain the pixel-level cost map $\mathcal{F}_{cv} $ as
\begin{equation}
    \mathcal{F}_{cv}(i,j,n,p)= \frac{\mathcal{F}_v(i,j) \cdot E(n,p)}{\lVert \mathcal{F}_v(i,j) \rVert \lVert E(n,p) \rVert} ,
\end{equation}
where $i,j$ indicate the 2D spatial position,  $n$ represents the index of text embeddings, and $p$ represents the index of templates. Therefore, the initial cost map $\mathcal{F}_{cv}$ has the size of $H_v \times W_v \times N  \times P$. 
The initial cost map goes through a  convolutional layer  to generate the input feature map $\mathcal{F}_{dec}^{l1} \in \R^{H_v \times W_v\times  N \times D}$ of the  decoder. For simplicity, we do not show $\mathcal{F}_{dec}^{l1}$ in Fig. \ref{fig:main_arch}.

\subsection{Gradual Fusion Decoder}
Semantic segmentation greatly benefits from high-resolution feature maps. However, the cost map $\mathcal{F}_{cv}$ generated by encoder has a relatively low resolution and high noise. Therefore, it is not beneficial to generate high-quality segmentation map  by directly  using cost map for prediction. To address this issue, we propose a gradual fusion decoder (GFD). GFD gradually generates high-resolution feature map $\mathcal{F}_h$ by cascading two modules, including feature aggregation module (\sta) and  skip-layer fusion module (\ffu), into multiple layers. \sta aims to model the relationship between local regions and different classes, whereas \ffu is designed to  enhance the local details of feature maps using shallow  features of hierarchical encoder.

\begin{figure}[t]
\centering
\includegraphics[width=1.0\linewidth]{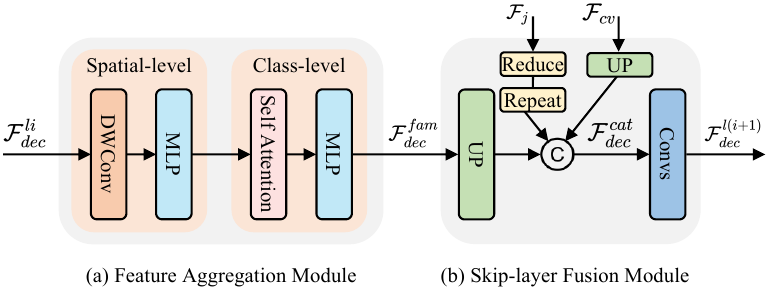}
\vspace{-10pt}
\caption{\textbf{Structure of gradual fusion decoder.} The gradual fusion decoder (GFD) first performs feature aggregation (a) in both spatial and class levels, and then employs skip-layer fusion (b) to combine the feature maps from previous decoder layer and  hierarchical encoder.}
\label{fig:gfd}
\vspace{-10pt}
\end{figure}

\noindent\textbf{Feature Aggregation Module:}
Fig. \ref{fig:gfd}(a) shows the design of the feature aggregation module (FAM) that has spatial-level  and class-level fusion. We first perform spatial-level fusion to model the relationship of local region. Prior works~\cite{convnext,peng2017large} have demonstrated that large-kernel convolutional operation is a simple but efficient structure to capture local information. Motivated by this, we perform spatial-level fusion employing large-kernel convolution \cite{convnext}. Specifically, the input feature map $\mathcal{F}_{dec}^{li}$  goes through a depth-wise convolutional layer and an MLP layer. The depth-wise convolutional layer has a $9\times9$ depth-wise convolution and a layer-norm operation, and the MLP layer contains two linear layers and a GeLU layer. In addition, we use a residual connection in both convolutional and MLP layers. Following the spatial-level aggregation, we further apply a linear self-attention operation as in \cite{catseg,katharopoulos2020transformers} along  category dimension to perform class-level feature aggregation. The generated  feature map by feature aggregation module (FAM) is represented as $\mathcal{F}_{dec}^{fam}$.

\noindent\textbf{Skip-layer Fusion Module:}
The  feature map $\mathcal{F}_{dec}^{fam}$ is spatially coarser, which lacks local detail information. In contrast, the shallow feature maps $\mathcal{F}_{2},\mathcal{F}_{3},\mathcal{F}_{4}$ in hierarchical encoder contains rich detail information. 
To incorporate these local details for segmentation, we introduce the skip-layer fusion module to gradually combine the low-resolution feature map $\mathcal{F}_{dec}^{fam}$  with high-resolution feature maps  $\mathcal{F}_{2},\mathcal{F}_{3},\mathcal{F}_{4}$. As shown in Fig. \ref{fig:gfd}(b), we first upsample low-resolution feature map  $\mathcal{F}_{dec}^{fam}$  by a factor of $2$  using the deconvolutional operation. Then, we reduce the channel dimension of the corresponding high-resolution feature map $\mathcal{F}_{j}, j\in 2,3,4$ by a factor of 16 using the convolutional operation, and repeat the reduced feature map $N$ times to have the same category dimension with $\mathcal{F}_{dec}^{fam}$. Afterwards, we  concatenate the upsampled feature map and the repeated feature map together. To fuse more information, we also upsample and concatenate the initial cost map $\mathcal{F}_{cv}$. Finally, we feed the concatenated feature map $\mathcal{F}_{dec}^{cat}$ through two convolutional layers to generate the output feature map $\mathcal{F}_{dec}^{l(i+1)}$.  As observed in \cite{catseg}, directly back-propagating the gradient to the image encoder degrades the  performance of open-vocabulary semantic segmentation. Therefore, we stop gradient back-propagation directly from skip-layer fusion module to the image encoder. 

Our observation (see Table \ref{tab:skip_layer}) reveals that, compared to plain transformer, hierarchical encoder with skip-layer fusion significantly improves the performance. This is likely due to that ,the hierarchical encoder is able to provide rich local information for segmentation, and the stopped gradient back-propagation avoids the negative impact on open-vocabulary segmentation ability. 

\begin{figure}[t]
\centering
\includegraphics[width=1.0\linewidth]{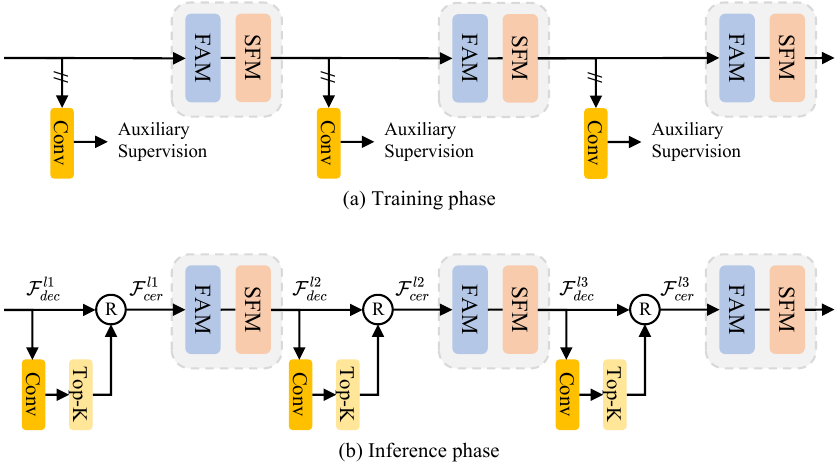}
\vspace{-10pt}
\caption{\textbf{Structure of category early rejection.} During training (a), we attach an auxiliary convolution after each decoder layer to predict segmentation maps  supervised by ground-truths. During inference (b), we employ top-$k$  strategy to predict existing categories and reject non-existing categories for next decoder layer.}
\vspace{-10pt}
\label{fig:CERS}
\end{figure}

\subsection{Category Early Rejection}
The computational cost of gradual fusion decoder is proportional to the number of semantic categories. When the number of categories is very large, the inference time significantly increases. In fact, most images only  contain several  semantic categories. As a result, majority of the inference time of the decoder is taken to calculate the features of the  non-existing categories. To boost the inference speed, we introduce a category early rejection scheme to recognize these existing categories and reject non-existing categories at the early decoder layer. The feature maps corresponding to rejected categories are removed from current decoder layer, and the following decoder layer only considers the reserved categories.

During training, as shown in Fig. \ref{fig:CERS}(a),  we add the auxiliary convolutional branch after each layer to respectively predict  segmentation maps, which are supervised by ground-truths. 
 To avoid the negative effect on model training, we stop their gradient back-propagation  to the decoder. 
 
During inference, we employ a top-$k$ strategy on segmentation maps to predict the existing  semantic categories. Specifically, we select the top-$k$ categories with maximum responses for each pixel and  generate a union set of categories from all pixels, which is fed to next decoder layer. We observe that $k=8$ can   ensure that most existing categories is recognized.
Fig. \ref{fig:CERS}(b) shows the category early rejection during inference. We first predict segmentation maps from $\mathcal{F}_{dec}^{l1}$ and employ the top-$k$ strategy to select $N_{l1}$ categories. Then, we remove the feature maps of non-selected categories and generate the output feature map $\mathcal{F}_{cer}^{l1} \in \mathbb{R}^{H_{v} \times W_{v} \times N_{l1} \times D}$. The generated feature map $\mathcal{F}_{cer}^{l1}$ is fed to the decoder layer.  Similarly, we generate the feature maps with fewer categories for the following layers. Therefore, most non-existing categories are rejected at early layer, which boosts the inference speed of the decoder.

\section{Experiments}\label{experiment}
\subsection{Datasets and Evaluation Metric}
Following existing open-vocabulary semantic segmentation methods~\cite{catseg,san}, we use the large-scale dataset COCO-Stuff~\cite{caesar2018coco} to train the model. In COCO-Stuff dataset, the training set contains about 118$k$ densely-annotated images with 171 different semantic categories. With the trained model on COCO-Stuff, we conduct experiments on multiple widely-used semantic segmentation datasets (ADE20K \cite{ade20k}, PASCAL VOC \cite{everingham2010pascal}, and PASCAL-Context \cite{pascal_context}) to demonstrate the effectiveness of the proposed SED and compare it with state-of-the-art methods in literature.

\noindent\textbf{ADE20K \cite{ade20k}}  is a large-scale semantic segmentation dataset. It contains $20k$ training images and $2k$ validation images. In open-vocabulary semantic segmentation task, there are two different test sets: \texttt{A-150} and \texttt{A-847}. The test set \texttt{A-150} has 150 common categories, while the test set \texttt{A-847} has 847 categories. 

\noindent\textbf{PASCAL VOC \cite{everingham2010pascal}} is one of the earliest datasets for object detection and segmentation. There are about $1.5k$ training images and $1.5k$ validation images. The dataset contains 20 different object categories. In open-vocabulary semantic segmentation task, we name it as \texttt{PAS-20}.

\noindent\textbf{PASCAL-Context \cite{pascal_context}} is extended from the original PASCAL VOC dataset  for semantic segmentation. In open-vocabulary semantic segmentation, there are two different test sets: \texttt{PC-59} and \texttt{PC-459}. The test set \texttt{PC-59} has 59  categories, while the test set \texttt{PC-459} has 459 categories.

\noindent\textbf{Evaluation metric:} Following existing traditional and open-vocabulary semantic segmentation, we adopt mean Intersection over Union (mIoU) as  evaluation metric. It is the averaged value of   intersection over unions over all classes. 

\begin{table*}[ht]
\footnotesize
    \begin{center}
    \resizebox{\textwidth}{!}{
    \begin{tabular}{l|cc|c|ccccc}
    \toprule
        Method & VLM & Feature backbone & Training dataset & \texttt{A-847} & \texttt{PC-459} & \texttt{A-150} & \texttt{PC-59} & \texttt{PAS-20} 
        \\
        \midrule
        SPNet~\cite{SPNet} & - & ResNet-101 & PASCAL VOC & - & - & - & 24.3 & 18.3  \\
        ZS3Net~\cite{ZS3Net} & - & ResNet-101 & PASCAL VOC & - & - & - & 19.4 & 38.3 \\
        LSeg~\cite{lseg} & ViT-B/32 & ResNet-101 & PASCAL VOC-15  & - & - & - & - & 47.4 \\
        LSeg+~\cite{openseg} & ALIGN & ResNet-101 & COCO-Stuff & 2.5 & 5.2 & 13.0 & 36.0 & -  \\
        Han et al.~\cite{han2023global} & ViT-B/16 & ResNet-101& COCO Panoptic~\cite{kirillov2019panoptic} & 3.5 & 7.1 & 18.8 & 45.2 & 83.2  \\
        GroupViT~\cite{groupvit} & ViT-S/16 & - & GCC~\cite{sharma2018conceptual}+YFCC~\cite{thomee2016yfcc100m} & 4.3 & 4.9 & 10.6 & 25.9 & 50.7   \\
        ZegFormer~\cite{zegformer} & ViT-B/16 & ResNet-101 & COCO-Stuff-156 & 4.9 & 9.1 & 16.9 & 42.8 & 86.2 \\
        ZegFormer~\cite{catseg} & ViT-B/16 & ResNet-101 & COCO-Stuff & 5.6 & 10.4 & 18.0 & 45.5 & 89.5  \\
        SimBaseline~\cite{xu2022simple} & ViT-B/16 & ResNet-101 & COCO-Stuff & 7.0 & - & 20.5 & 47.7 & 88.4\\
        OpenSeg~\cite{openseg} & ALIGN & ResNet-101 & COCO Panoptic~\cite{kirillov2019panoptic}+LOc. Narr.~\cite{pont2020connecting} & 4.4 & 7.9 & 17.5 & 40.1 & - \\
        DeOP~\cite{Han2023ZeroShotSS} & ViT-B/16 & ResNet-101c& COCO-Stuff-156  & 7.1 & 9.4 & 22.9 & 48.8 & 91.7  \\
        PACL~\cite{mukhoti2023open} & ViT-B/16 & - & 
        GCC~\cite{sharma2018conceptual}+YFCC~\cite{thomee2016yfcc100m}  & - & - & \underline{31.4} & 50.1 & 72.3  \\
        OVSeg~\cite{ovseg} & ViT-B/16 & ResNet-101c & COCO-Stuff+COCO Caption & 7.1 & 11.0 & 24.8 & 53.3 & 92.6  \\
        CAT-Seg~\cite{catseg} & ViT-B/16 & ResNet-101 & COCO-Stuff & 8.4 & \underline{16.6} & 27.2 & \textbf{57.5} & 93.7 \\
        SAN~\cite{san} & ViT-B/16 & -&  COCO-Stuff & \underline{10.1} & 12.6 & 27.5 & 53.8 & \underline{94.0}   \\
        \hlrow 
        \hlrow 
        \textbf{\modelname (Ours)} & ConvNeXt-B & - & COCO-Stuff & \textbf{11.4}  &\textbf{18.6} &\textbf{31.6}  &\underline{57.3} &  \textbf{94.4}\\
        \midrule
        LSeg~\cite{lseg} & ViT-B/32 & ViT-L/16 & PASCAL VOC-15  & - & - & - & - & 52.3  \\
        OpenSeg~\cite{openseg} & ALIGN & Eff-B7~\cite{tan2019efficientnet} & COCO Panoptic~\cite{kirillov2019panoptic}+LOc. Narr.~\cite{pont2020connecting} & 8.1 & 11.5 & 26.4 & 44.8 & - \\
        OVSeg~\cite{ovseg} & ViT-L/14 & Swin-B & COCO-Stuff+COCO Caption & 9.0 & 12.4 & 29.6 & 55.7 & 94.5  \\
        Ding \textit{et al.}~\cite{maskclip} & ViT-L/14 & - & COCO Panoptic~\cite{kirillov2019panoptic} & 8.2 & 10.0 & 23.7 & 45.9 & -  \\
        ODISE~\cite{odise} & ViT-L/14 & - & COCO Panoptic~\cite{kirillov2019panoptic}& 11.1 & 14.5 & 29.9 & 57.3 & - \\
        HIPIE~\cite{wang2023hierarchical} & BERT-B~\cite{devlin2018bert} & ViT-H & COCO Panoptic~\cite{kirillov2019panoptic}& - & - & 29.0 & 59.3 & -\\
        SAN~\cite{san} & ViT-L/14 & -&  COCO-Stuff & 13.7 & 17.1 & 33.3 & 60.2 & 95.5  \\
        CAT-Seg~\cite{catseg} & ViT-L/14 & Swin-B & COCO-Stuff & 10.8 & \underline{20.4} & 31.5 & \textbf{62.0} & \textbf{96.6} \\
        FC-CLIP~\cite{fcclip} & ConvNeXt-L & - & COCO Panoptic~\cite{kirillov2019panoptic} & \textbf{14.8} & 18.2 & \underline{34.1} & 58.4 & 95.4  \\
        \hlrow 
        \hlrow 
        \textbf{\modelname (Ours)} & ConvNeXt-L & - & COCO-Stuff & \underline{13.9} & \textbf{22.6} & \textbf{35.2} & \underline{60.6} & \underline{96.1}  \\
        \bottomrule
    \end{tabular}
    }
    \vspace{-5pt}
    \caption{\textbf{Comparison with state-of-the-art methods.} We report the mIoU results  on five widely used  test sets for open-vocabulary semantic segmentation. Here, the best results are shown in bold and the second-best results are underlined. With comparable VLM model, Our proposed SED method achieves superior performance on all five test sets. Compared to CAT-Seg \cite{catseg} and OVSeg \cite{ovseg}, our proposed SED method does not use additional backbone or dataset.}\label{tab:main_table}
    \end{center}
    \vspace{-10pt}
\end{table*}

\begin{table}[t]
\scriptsize
\centering
\subfloat[With base model]{
\begin{tabular}{l|cc}
\toprule
Method  &  mIoU & Time \\
\midrule
        SimBaseline~\cite{xu2022simple} & 20.5 & 316  \\
        OVSeg~\cite{ovseg} & 24.8 & 314  \\
        CAT-Seg~\cite{catseg} & 27.2 & 362  \\
        SAN~\cite{san} & 27.5 & \textbf{32}  \\
        % \textbf{\modelname (ours)}& 31.6 & 81 \\
        \hlrow \textbf{\modelname (ours)}& \textbf{31.6} & 82 \\
        \hlrow \textbf{\modelname-fast (ours)} & 29.4 & \textbf{32} \\
\bottomrule
\end{tabular}
}
\subfloat[With large model]{
\begin{tabular}{l|cc}
\toprule
Method  & mIoU & Time  \\
\midrule
        % OVSeg~\cite{ovseg} & 29.6 & 1230 \\
        ODISE~\cite{odise} & 29.9 & 1989 \\
        CAT-Seg~\cite{catseg} & 31.5 & 433 \\
        SAN~\cite{san} & 33.3 & 117 \\
        FC-CLIP~\cite{fcclip} & 34.1 & 285 \\
        \hlrow \textbf{\modelname (ours)} & \textbf{35.2} & 98 \\
        \hlrow \textbf{\modelname-fast (ours)} & 34.2 & \textbf{64} \\
\bottomrule
\end{tabular}
}
\vspace{-5pt}
\caption{\textbf{Comparison in terms of mIoU and inference time (ms).} We report the results on \texttt{A-150} with base and large models. Here, the inference time is reported on a single NVIDIA A6000 GPU.}
\label{tab:baselarge}
\vspace{-10pt}
\end{table}

\subsection{Implementation Details}

We adopt the pre-trained vision-language model CLIP~\cite{schuhmann2022laion,cherti2023reproducible,CLIP} as the base model, where the hierarchical backbone ConvNeXt-B or ConvNeXt-L is used as hierarchical image encoder. The feature dimension $D_t$ of text embeddings are 640 for ConvNeXt-B and 768 for ConvNeXt-L,  the number of category templates $P$ is 80, and the channel number of  feature map $F_{dec}^{l1}$ is 128. We freeze the text encoder and only train the image encoder and gradual fusion decoder. We train our model on 4 NVIDIA A6000 GPUs with the mini-batch of 4 images. The optimizer AdamW is adopted with the initial learning rate of 2$\times10^{-4}$ and the weight decay of 1$\times10^{-4}$. To avoid over-fitting on training set, the learning rate of image encoder is multiplied by a factor of $\lambda = 0.01$. There are totally 80$k$ training iterations. During training, we crop the input image with the $768\times768$ pixels. During inference, the input image is resized with the $768\times768$ pixels.  

\subsection{Comparisons With State-of-the-art Methods}
Here, we compare our proposed  SED with some state-of-the-art open-vocabulary semantic segmentation methods. Table \ref{tab:main_table} presents the results of different methods on all five test sets. It also shows the corresponding vision-language model (VLM), feature backbone, and training dataset. Most methods, except SPNet \cite{SPNet} and ZS3Net \cite{ZS3Net}, are developed based on VLM. Some methods \cite{ovseg,catseg} employ additional  feature backbones, while some methods \cite{openseg,ovseg} use additional dataset or annotation.
\begin{table}[t]
    \centering
    \resizebox{\linewidth}{!}{
   \begin{tabular}{ccc|ccccc}
        \toprule
         HECG & GFD & CER & \texttt{A-847} & \texttt{PC-459} & \texttt{A-150} & \texttt{PC-59} & \texttt{PAS-20} 
         \\
        \midrule
        & & & 7.3& 14.9& 23.7& 52.9&94.4\\
        \checkmark & &  & 9.9 & 17.2 & 28.2 & 54.7 & 95.0 \\
        \checkmark & \checkmark & & 11.2& 18.6& 31.8& 57.7& 94.4 \\
        \checkmark & \checkmark & \checkmark & 11.4& 18.6& 31.6& 57.3& 94.4\\
        \bottomrule
\end{tabular}}%
\vspace{-5pt}
        \caption{\textbf{Impact of different modules in our \modelname.} We show the results of integrating different modules into the baseline.}
    \label{tab:main_components}
    \vspace{-10pt}
\end{table}

Most existing open-vocabulary semantic segmentation methods are developed on VLM with plain transformer ViT, including two-stage  OVSeg \cite{ovseg} and single-stage  CAT-Seg \cite{catseg} and SAN \cite{san}. In contrast, our proposed  SED adopts hierarchical encoder ConvNeXt. When using the comparable image encoder ViT-B or ConvNeXt-B, our  SED outperforms these methods on all five test sets. On \texttt{PC-459}, our SED outperforms OVSeg \cite{ovseg}, CAT-Seg \cite{catseg},  and SAN \cite{san} by 7.6\%, 2.0\%, and 6.0\%. On \texttt{A-150}, our SED outperforms OpenSeg \cite{openseg}, CAT-Seg \cite{catseg},  and SAN \cite{san} by 14.1\%, 4.4\%, and 4.1\%. Moreover, compared to OVSeg and CAT-Seg, our SED does not require additional feature backbone. Compared to OVSeg and OpenSeg, our SED does not require additional dataset or annotation.

When using the comparable image encoder ViT-L or ConvNeXt-L, our  SED also achieves favourable performance  on all five test sets. For example, on \texttt{PC-459}, our SED outperforms  SAN \cite{san}, CAT-Seg \cite{catseg},  and FC-CLIP \cite{fcclip} by 5.5\%, 2.2\%, and 4.4\%. 

Table \ref{tab:baselarge} further shows the accuracy and speed comparison on \texttt{A-150}. Compared to most methods, our proposed SED has the results with both base and large models at fast speed. We also present a faster version (SED-fast) by downsampling the input size.  Compared to SAN, our SED-fast is 1.9\% better at similar speed with base model, and is 0.9\% better and 1.8 times faster with large model.

\begin{table}[t]
\centering
\resizebox{\linewidth}{!}{
\begin{tabular}{l|c|ccccc}
    \toprule
    Image Encoder& Skip-layer& \texttt{A-847} & \texttt{PC-459} & \texttt{A-150} & \texttt{PC-59} & \texttt{PAS-20} \\
    \midrule
    \multirow{2}{*}{Plain} & w/o & 7.3& 13.5& 23.0& 51.5& 94.1\\
  &  with & 7.3& 14.9& 23.7& 52.9& 94.4\\
    \midrule
  \multirow{2}{*}{Hierarchical} & w/o& 7.9& 14.3 & 25.7& 52.0& 92.7\\
  &  with & 9.9& 17.2& 28.2& 54.7& 95.0\\
    \bottomrule
\end{tabular}}
\vspace{-5pt}
    \caption{\textbf{Comparison of plain and hierarchical encoder.} We employ ViT-B and ConvNeXt-B as plain encoder and hierarchical encoder, respectively.}
\label{tab:skip_layer}
\vspace{-10pt}
\end{table}  
\subsection{Ablation Study}
Here we perform ablation study to show the efficacy of our proposed method using ConvNeXt-B as  image encoder.

\noindent\textbf{Impact of integrating different components:} Table \ref{tab:main_components} shows the impact of integrating different components  into the baseline. The baseline adopts original CLIP with plain transformer ViT and uses the cost map to predict segmentation map with skip-layer fusion. The baseline obtains the mIoU scores of 7.3\%, 14.9\%, and 23.7\% on \texttt{A-847}, \texttt{PC-459}, and \texttt{A-150}. When using hierarchical encoder to replace plain transformer, it has the mIoU scores of 9.9\%, 17.2\%, and 28.2\% on \texttt{A-847}, \texttt{PC-459}, and \texttt{A-150}, which outperforms the baseline by 2.6\%, 2.3\%, and 4.5\%. When further integrating gradual fusion decoder, it has the mIoU scores of 11.2\%, 18.6\%, and 31.8\% on \texttt{A-847}, \texttt{PC-459}, and \texttt{A-150}, which outperforms the baseline by 3.9\%, 3.7\%, and 8.1\%. When further integrating category early rejection strategy into our method, it  almost does not degrade  performance but has a faster speed (see Table \ref{tab:topk}).

\begin{table}[t]
    \centering
    \resizebox{\linewidth}{!}{
   \begin{tabular}{l|c|ccccc}
        \toprule
        \multirow{6}{*}{(a)}  & Strategy & \texttt{A-847} & \texttt{PC-459} & \texttt{A-150} & \texttt{PC-59} & \texttt{PAS-20}
        \\
        \cmidrule{2-7}
       & Freeze All & 9.4& 15.3& 28.6& 49.7& 77.2\\
      &  Freeze L0-L2 & 11.2& 18.3& 30.6& 54.9& 91.5\\
      &  Freeze L0-L1 & 10.4 & 17.5 & 31.4 & 57.3 & 94.0 \\
      &  Freeze L0 & 10.6& 17.7& 31.6& 57.2& 94.1\\
      &  Fine-tune All & 11.2& 18.6& 31.8& 57.7& 94.4\\
        \midrule
        % \midrule
      \multirow{4}{*}{(b)} &  Factor $\lambda$ & \texttt{A-847} & \texttt{PC-459} & \texttt{A-150} & \texttt{PC-59} & \texttt{PAS-20} 
         \\
        \cmidrule{2-7}
      &   0.005& 11.3& 17.6 & 31.6& 56.4& 93.6\\
      &  0.01 & 11.2& 18.6& 31.8& 57.7& 94.4\\
      &   0.02 & 10.5 & 17.7& 31.3& 57.7& 94.7\\
        \bottomrule
\end{tabular}}\vspace{-5pt}
        \caption{\textbf{Ablation study on fine-tuning image encoder in HECG.} We show the results of different fine-tuning strategies and different scale factors of encoder learning rates.}
    \label{tab:enc}
\vspace{-10pt}
\end{table}

\begin{table}[t]
\centering
\resizebox{\linewidth}{!}{
\begin{tabular}{c|c|cccccc}
    \toprule
  \multirow{4}{*}{(a)} & Kernel Size & \texttt{A-847} & \texttt{PC-459} & \texttt{A-150} & \texttt{PC-59} & \texttt{PAS-20} 
    \\
    \cmidrule{2-7}
   & 7& 11.1& 18.0& 31.8& 57.3& 93.9\\
   & 9 & 11.2& 18.6& 31.8& 57.7& 94.4\\
   & 11 & 10.8& 18.0& 31.8& 57.1& 94.5\\
    \midrule
    % \midrule
 \multirow{4}{*}{(b)} &  Aggregation & \texttt{A-847} & \texttt{PC-459} & \texttt{A-150} & \texttt{PC-59} & \texttt{PAS-20} 
     \\
    \cmidrule{2-7}
  &  Spatial-level & 10.1& 17.6& 28.8& 54.8& 94.4\\
  &  Class-level & 10.0& 17.4& 30.7& 56.0& 92.7\\
  &  Both & 11.2& 18.6& 31.8& 57.7& 94.4\\
    \midrule  
     % \midrule
 \multirow{4}{*}{(c)} &  Feature Fusion & \texttt{A-847} & \texttt{PC-459} & \texttt{A-150} & \texttt{PC-59} & \texttt{PAS-20} 
     \\
    \cmidrule{2-7}
  &  None & 7.9 & 14.3 & 25.7 & 52.0 & 92.7\\
  &  +$\mathcal{F}_{2,3,4}$ & 11.1 & 17.9 & 32.0 & 57.5 & 94.2\\
  &  +$\mathcal{F}_{2,3,4}$+$\mathcal{F}_{cv}$ & 11.2& 18.6& 31.8& 57.7& 94.4\\
    \midrule    
    % \midrule
 \multirow{4}{*}{(d)} &  Gradient to $\mathcal{F}_{2,3,4}$  & \texttt{A-847} & \texttt{PC-459} & \texttt{A-150} & \texttt{PC-59} & \texttt{PAS-20}
     \\
    \cmidrule{2-7}
  &  w/o Stop & 10.6 & 18.0 & 31.7 & 57.3 & 93.9\\
  &  with Stop & 11.2& 18.6& 31.8& 57.7& 94.4\\
    \midrule      
  \multirow{4}{*}{(e)} & Decoder Layer & \texttt{A-847} & \texttt{PC-459} & \texttt{A-150} & \texttt{PC-59} & \texttt{PAS-20}
     \\
    \cmidrule{2-7}
   & 1 & 10.1 & 17.0& 29.8& 55.6& 91.0\\
  &  2 & 11.1& 18.3& 31.8& 57.3& 93.8\\
  &  3 & 11.2& 18.6& 31.8& 57.7& 94.4\\
    \bottomrule
\end{tabular}}
\vspace{-5pt}
    \caption{\textbf{Ablation study on different designs in GFD.} Gradually fusion decoder (GFD) contains feature aggregation module (FAM) and skip-layer fusion module (SFM). We first show the impact of different kernel sizes (a) and spatial-class aggregation (b) in FAM.  Then, we give the impact of fusing different feature maps (c) and gradient back-propagation (d) in SFM. Finally, we show the impact of different decoder layers (e).}
\label{tab:gfd}
\vspace{-10pt}
\end{table}

\noindent\textbf{Plain \textit{vs} hierarchical encoder:} Table \ref{tab:skip_layer} compares plain ViT-B or hierarchical ConvNeXt-B used as image encoder. Hierarchical encoder outperforms plain encoder with or without using skip-layer connection. When using skip-layer connection, hierarchical encoder has a larger improvement. Therefore, different feature maps of hierarchical encoder   provide rich local information for segmentation.

\noindent\textbf{Ablation study on fine-tuning encoder in HECG:} Table \ref{tab:enc} presents  ablation study on fine-tuning   hierarchical encoder in HECG. In top part (a), we show the impact of different fine-tuning strategies. When we freeze all the layers in the encoder, it has the lowest performance on all five test sets. When we fine-tune all the layers in the encoder, it achieves the best results on all test sets. In bottom part (b), we present the impact of different scale factor $\lambda$ of encoder learning rates. With the scale factor is 1$\times 10^{-2}$, it has the best performance. A larger or smaller scale factor will degrade the performance in some degree. Therefore, we fine-tune hierarchical encoder using the scale factor of 1$\times 10^{-2}$.

\begin{figure*}[t]
\centering
\includegraphics[width=0.96\linewidth]{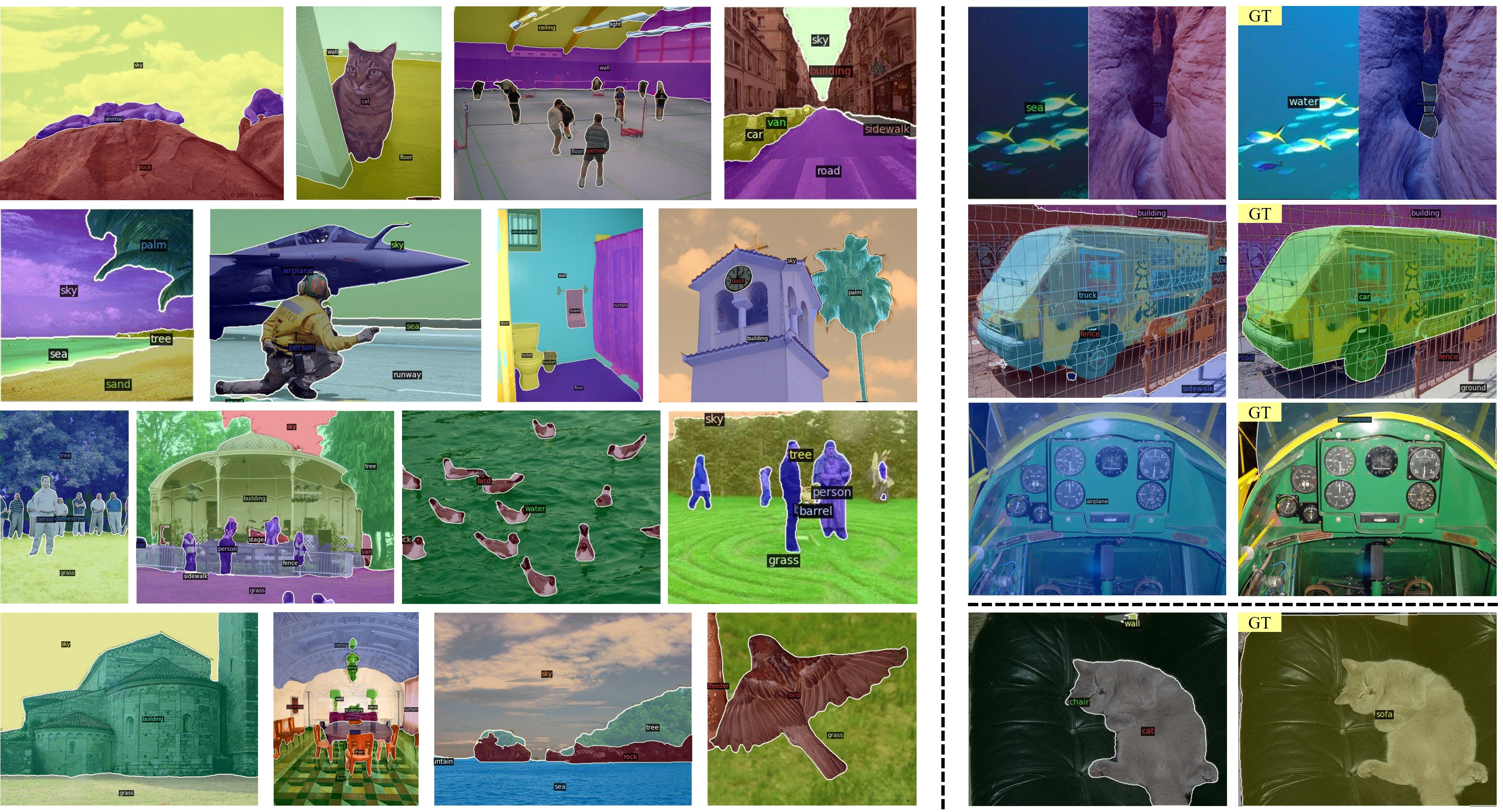}
\vspace{-5pt}
\caption{\textbf{Qualitative results.} In the left part, we show some high-quality results, where our  method can accurately classify and segment various categories. In the right-top part, we give some failure cases and corresponding ground-truths (GT). In the right-bottom part, we give one case in which our method can segment the cat that is not annotated in ground-truths (GT).}
\label{fig:qualitative}
\vspace{-10pt}
\end{figure*}

\noindent\textbf{Ablation study on GFD:} Table \ref{tab:gfd} presents ablation study on different designs in gradual fusion decoder (GFD). Our gradual fusion decoder contains feature aggregation module (FAM) and skip-layer fusion module (SFM). We first perform some experiments on FAM. In (a), we show the impact of different large-kernel sizes. It has the best results using the kernel size of 9. We also observe that large-kernel operation is better and faster than Swin block. On PC-459, the mIoU has  0.4\% improvement and the speed is 1.2 times faster. In (b), we present the impact of spatial-level and class-level feature aggregation. Both spatial-level and class-level feature aggregation  improve performance on five test sets. When combining them together, it has the best results. 

Afterwards, we perform some experiments on SFM. In (c), we present the impact of fusing different feature maps in SFM. It can be seen that integrating different feature maps can significantly improve performance. In (d), we present the impact of gradient back-propagation in SFM. It has better performance without gradient back-propagation. Finally, we show the impact of different decoder layers in (e). Compared to using only one decoder layer, using three decoder layers has the best results. For example, it has 1.6\% improvement on \texttt{PC-459}.

\begin{table}[t]
    \centering
    \resizebox{\linewidth}{!}{%
    \begin{tabular}{l|c|cccccc}
    \toprule
       Metric  & $k$ & \texttt{A-847} & \texttt{PC-459} & \texttt{A-150} & \texttt{PC-59} & \texttt{PAS-20}
        \\
        \midrule
        \multirow{5}{*}{mIoU} & 1 & 10.2& 17.6& 29.9& 55.9& 93.8\\
         & 2 & 10.9& 18.3& 30.8& 56.7& 94.0\\
         & 4 & 11.2& 18.5& 31.4& 57.0& 94.2\\
         & 8 & 11.4& 18.6& 31.6& 57.3& 94.4\\
         & All & 11.2& 18.6& 31.8& 57.7& 94.4\\
        \midrule
        \multirow{5}{*}{Time (ms)} & 1 & 120.8& 74.8& 51.3& 40.5& 37.2\\
         & 2 & 136.4& 84.2& 55.8& 44.2& 38.9\\
         & 4 & 151.5& 93.4& 67.5& 49.7& 42.2\\
         & 8 & 181.6& 120.1& 82.4& 59.8& 47.3\\
         & All & 861.0& 468.1& 177.7& 84.7& 44.9\\       
        \bottomrule
    \end{tabular}%
    }
    \vspace{-5pt}
    \caption{\textbf{Impact of selecting top-$k$ categories in CER.} We show both mIoU and inference time (ms). 
    Here, the inference time is reported on a single NVIDIA A6000 GPU.
    }
    \label{tab:topk}
    \vspace{-10pt}
\end{table}

\noindent\textbf{Ablation study on CER:} Table \ref{tab:topk} presents the impact of selecting top-$k$ categories in category early rejection (CER). The small number of $k$ indicates selecting few categories to fed the  decoder layer, which can  accelerate inference speed but may sacrifice accuracy. For example, on \texttt{A-847},  when $k$ is equal to 1, the speed is 7.1 times faster than using all categories, but the mIoU is 1.0\% lower. When $k=8$, we observed a slight improvement in performance, with a speed increase of about 4.7 times.\\
\noindent\textbf{Qualitative results:} Fig. \ref{fig:qualitative} presents some qualitative results. The left part shows high-quality segmentation results. Our  method is able to accurately segment various categories, such as palm, runway, and sand. The right-top part shows some failure cases. In first two rows, our  method  mistakenly recognizes the water as sea, the earth as rock, and the car as truck. In third row, our method  mistakenly recognizes airplane and ignores windowpane.   In addition,  The right-bottom part shows that our  method successfully segments cat that is ignored by the \texttt{PC-59} ground-truth.

\section{Conclusion}
We propose an approach, named SED, for open-vocabulary semantic segmentation. Our SED comprises hierarchical encoder-based cost map generation and gradual fusion decoder with category early rejection. We first employ hierarchical encoder to generate pixel-level image-text cost map. Based on generated cost map and different feature maps in hierarchical encoder, we employ gradual fusion decoder to generate high-resolution feature map for segmentation. To boost speed, we introduce a category early rejection scheme into decoder to early reject non-existing categories. Experiments on multiple datasets reveal the effectiveness of our method in terms of both accuracy and speed. \\
\noindent\textbf{Future work:}  
Our model sometimes struggle on recognizing near-synonym categories as classes. In future, we will explore designing category attention strategy or using large-scale fine-grained dataset to solve this challenge.
{
    \small
    \bibliographystyle{ieeenat_fullname}
    \bibliography{main}
}

% WARNING: do not forget to delete the supplementary pages from your submission 
% \input{sec/X_suppl}

\end{document}